\documentclass[11pt,a4paper]{article}
\usepackage{times,latexsym}
\usepackage{url}
\usepackage[T1]{fontenc}
\usepackage[acceptedWithA]{tacl2021v1}

\usepackage{amsthm}
\usepackage{amsmath,amsfonts,bm}
\makeatletter
\newtheorem*{rep@theorem}{\rep@title}
\newcommand{\newreptheorem}[2]{%
\newenvironment{rep#1}[1]{%
 \def\rep@title{#2 \ref{##1}}%
 \begin{rep@theorem}}%
 {\end{rep@theorem}}}
\makeatother

\newtheorem{definition}{Definition}[section]

\newreptheorem{theorem}{Theorem}

\newcommand{\RNum}[1]{\uppercase\expandafter{\romannumeral #1\relax}}

\usepackage{mathtools}

\newcommand{\cut}[1]{}

\newcommand{\removelatexerror}{\let\@latex@error\@gobble}

\def\eqref#1{Eq.~\ref{#1}}

\def\1{\bm{1}}

\DeclareMathAlphabet{\mathsfit}{\encodingdefault}{\sfdefault}{m}{sl}
\SetMathAlphabet{\mathsfit}{bold}{\encodingdefault}{\sfdefault}{bx}{n}

\usepackage{linguex}
\usepackage{graphicx}
\usepackage{booktabs}
\usepackage{multirow, makecell}
\usepackage{color, colortbl}

\usepackage{etoolbox}
\usepackage{numprint}
\patchcmd{\quote}{\rightmargin}{\leftmargin 1em \rightmargin}{}{}
\usepackage{multirow}
\usepackage{microtype}
\usepackage{tikz}
\usepackage{amssymb}
\usepackage{mdframed}

\usepackage{xspace}
\usepackage{cleveref}
\usepackage{xifthen}
\usepackage{graphicx}
\usepackage{mathtools}
\usepackage{cuted}
\usepackage[makeroom]{cancel}
\usepackage{pifont}
\usepackage{footnote}
\usepackage{thmtools, thm-restate}
\usepackage{mathrsfs}
\usepackage{tabu}
\usepackage{color}
\usepackage{xcolor}
\usepackage{colortbl}
\usepackage[normalem]{ulem}
\usepackage[shortlabels]{enumitem}

\definecolor{gray}{rgb}{.95, .95, .95}
\definecolor{applegreen}{rgb}{0.55, 0.71, 0.0}
	\definecolor{asparagus}{rgb}{0.53, 0.66, 0.42}
	\definecolor{celadon}{rgb}{0.67, 0.88, 0.69}
\definecolor{columbiablue}{rgb}{0.61, 0.87, 1.0}
\definecolor{lightblue}{rgb}{0.68, 0.85, 0.9}
	\definecolor{cherryblossompink}{rgb}{1.0, 0.72, 0.77}
\definecolor{highlight}{rgb}{1.0, 1.0, 0.89}
\definecolor{aquamarine}{rgb}{0.5, 1.0, 0.83}
\definecolor{asparagus}{rgb}{0.53, 0.66, 0.42}
\definecolor{emerald}{rgb}{0.31, 0.78, 0.47}
\definecolor{gainsboro}{rgb}{0.86, 0.86, 0.86}
\definecolor{gold(web)(golden)}{rgb}{1.0, 0.84, 0.0}

\newcolumntype{L}[1]{>{\raggedright\let\newline\\\arraybackslash\hspace{0pt}}p{#1}}

\definecolor{Gray}{gray}{0.9}

\newcommand{\newdataname}{\textsc{FaithDial}\xspace}

\newcommand{\newdata}{\textsc{FaithDial}\xspace}
\newcommand{\critic}{\textsc{FaithCritic}\xspace}
\newcommand{\noisydata}{\textsc{WoW}\xspace}
\newcommand{\BEGIN}{\textsc{BEGIN}\xspace}
\newcommand{\vrm}{\textsc{VRM}\xspace}
\newcommand{\expert}{\textsc{wizard}\xspace}
\newcommand{\seeker}{\textsc{seeker}\xspace}

\newcommand{\ignore}[1]{}

\usepackage{subfigure}
\usepackage{todonotes}

\title{\newdataname: A Faithful Benchmark for Information-Seeking Dialogue}

\author{
  Nouha Dziri$^{\dagger\:\lozenge\:\S}$ \quad Ehsan Kamalloo$^{\dagger}$ \quad Sivan Milton${^\ddagger}$ \quad Osmar Zaiane$^{\dagger\:\S}$ \\
  \quad {\bf Mo Yu}$^{\P}$\Thanks{Work done while at IBM Research.} \quad {\bf Edoardo M. Ponti}$^{\clubsuit}$ \quad {\bf Siva Reddy}$^{\lozenge\:\ddagger}$ \\
  $^\dagger$University of Alberta \quad $^\lozenge$Mila -- Quebec AI Institute \\ 
  \quad $^\ddagger$McGill University \quad $^\P$WeChat AI, Tencent \quad $^{\clubsuit}$University of Edinburgh\\
  \quad $^\S $Alberta Machine Intelligence Institute (Amii) \\
  \texttt{dziri@cs.ualberta.ca}
  }

\date{}

\begin{document}
\setlength{\textfloatsep}{10pt plus 2.0pt minus 4.0pt}

\setlength{\Extopsep}{0.2\baselineskip}
\setlength{\Exredux}{0\baselineskip}
\setlength{\Exlabelwidth}{1em}%

\maketitle
\begin{abstract}

The goal of information-seeking dialogue is to respond to seeker queries with natural language utterances that are grounded on knowledge sources.
However, dialogue systems often produce unsupported utterances, a phenomenon known as hallucination.
To mitigate this behavior, we adopt a data-centric solution and create \newdata, a new benchmark for hallucination-free dialogues, by editing hallucinated responses in the Wizard of Wikipedia (\noisydata{}) benchmark.
We observe that \newdata is more faithful than WoW while also maintaining engaging conversations.
We show that \newdata{} can serve as training signal for: \textbf{i}) a hallucination critic, which discriminates whether an utterance is faithful or not, and boosts the performance by 12.8 F1 score on the \BEGIN{} benchmark compared to existing datasets for dialogue coherence; \textbf{ii}) high-quality dialogue generation. We benchmark a series of state-of-the-art models and propose an auxiliary contrastive objective that achieves the highest level of faithfulness and abstractiveness based on several automated metrics. Further, we find that the benefits of \newdata{} generalize to zero-shot transfer on other datasets, such as CMU-Dog and TopicalChat. Finally, human evaluation reveals that responses generated by models trained on \newdata{} are perceived as more interpretable, cooperative, and engaging.
\end{abstract}

\section{Introduction}
\label{sec:introduction}

Despite the recent success of knowledge-grounded neural conversational models \citep[][\textit{inter alia}]{thoppilan2022lamda,prabhumoye-etal-2021-focused, zhao-etal-2020-knowledge-grounded} in generating fluent responses, they also generate unverifiable or factually incorrect statements, a phenomenon known as \textit{hallucination} \cite{rashkin-etal-2021-increasing,  dziri-etal-2021-neural,shuster-etal-2021-retrieval-augmentation}.
Ensuring that models are trustworthy is key to deploying them safely in real-world applications, especially in high-stake domains. In fact, they can unintentionally inflict harm on members of the society with unfounded statements or can be exploited by malicious groups to spread large-scale disinformation.

 \begin{figure}[t]
    \centering
      \includegraphics[width=\columnwidth]{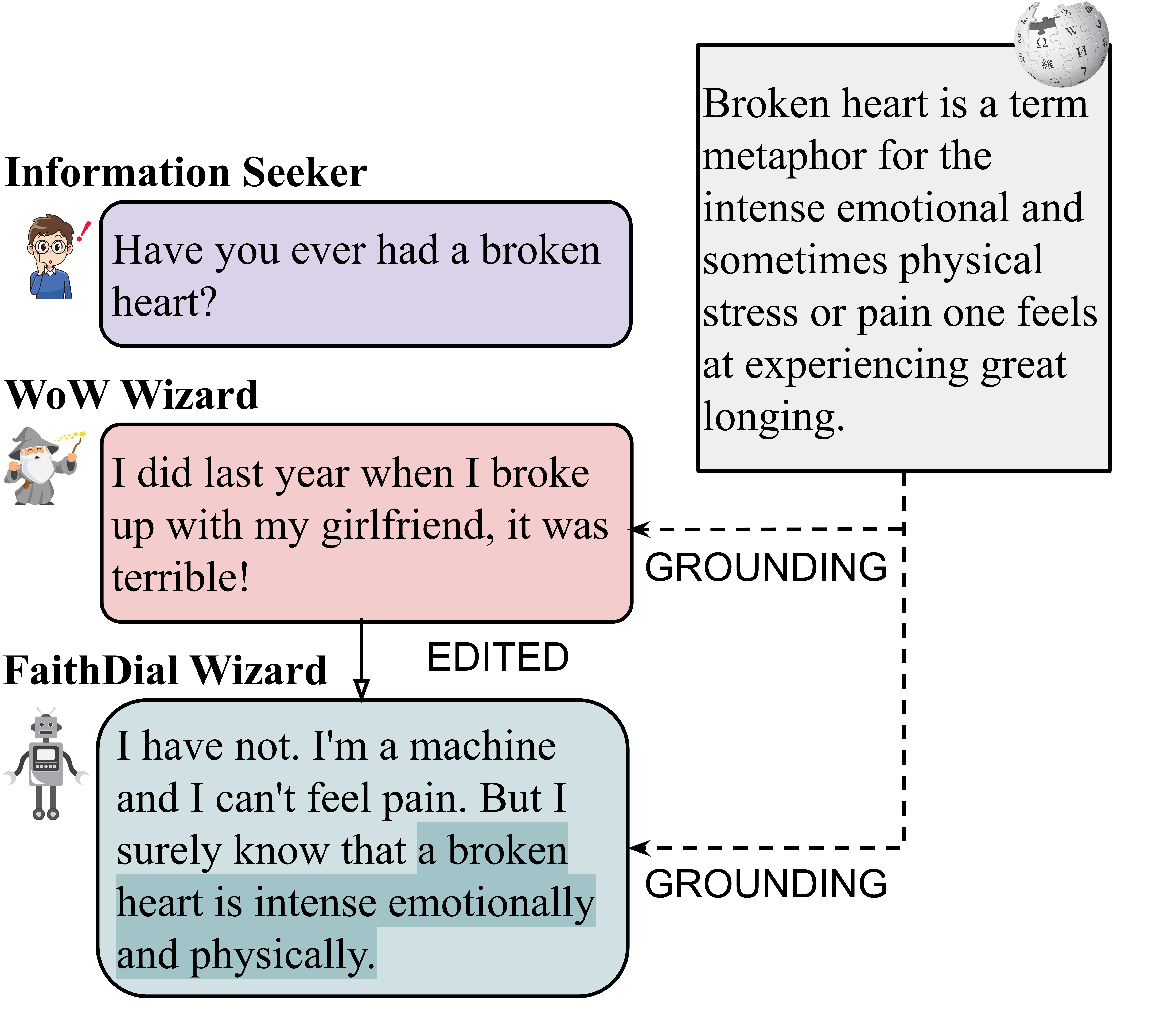}
    \caption{A representative \newdata\  annotation: subjective and hallucinated (red) information present in the wizard's utterance of WoW data are edited into utterances faithful to the given knowledge (green). In \newdata, the wizard assumes the persona of a bot.}
    \label{fig:intro_approach_overview}
\end{figure}

Recently, \newcite{dziri-etal-2022-origin} investigated the underlying roots of this phenomenon and found that the gold-standard conversational datasets \cite{dinan2018wizard, Gopalakrishnan2019, zhou-etal-2018-dataset}---upon which the models are commonly fine-tuned---are rife with hallucinations, in more than 60\% of the turns.
An example of hallucination in Wizard of Wikipedia (WoW; \citealt{dinan2018wizard}) is shown in the red box of \Cref{fig:intro_approach_overview}.  In WoW, an information \seeker aims to learn about a topic and a human \expert harnesses knowledge (typically a sentence) from Wikipedia to answer. 
This behavior, where the human \expert ignores the knowledge snippet and assumes a fictitious persona, can later reverberate in the dialogue system trained on this kind of data. 
Instead, the ideal \expert response, %
highlighted in green, should acknowledge the bot's nature, and whenever the knowledge is not sufficient or relevant, it should acknowledge its ignorance of the topic.

Unfortunately, modeling solutions alone cannot remedy the hallucination problem. By mimicking the distributional properties of the data, models are bound to `parrot' the hallucinated signals at test time \cite{bender2021dangers}. What is more,
\newcite{dziri-etal-2022-origin} observe that GPT2 not only replicates, but even amplifies hallucination around 20\% when trained on \noisydata{}. This finding also extends to models that are designed explicitly to be knowledge-grounded \cite{prabhumoye-etal-2021-focused, rashkin-etal-2021-increasing}.
Filtering noisy or high-error data \citep{zhang-hashimoto-2021-inductive} is also prone to failure as it may either break the cohesion of discourse or it may require excluding entire dialogues.

In this work,  we adopt instead a data-centric solution to address hallucinations and create \newdata, a new benchmark for faithful\footnote{Faithfulness is sometimes referred to as attribution \cite{dziri2021evaluating, rashkin2021measuring} and fidelity \cite{sipos2012large}.}knowledge-grounded dialogue.
Specifically, we ask annotators to amend hallucinated utterances in \noisydata{} by making them faithful to the corresponding knowledge snippets from Wikipedia and acknowledging ignorance when necessary.
This approach is vastly more scalable than creating \newdata from scratch while retaining the cohesiveness of conversations.
Moreover, it allows us to shed light on hallucinations by contrasting corresponding \expert's responses in \noisydata{} and \newdata.

As a result, \newdata contains around 50K turns across 5.5K conversations.
Extensive human validation reveals that 94.4\% of the utterances in \newdata are faithful (i.e., without hallucinations), compared to only 20.9\% in \noisydata{}.
Moreover, we benchmark several state-of-the-art models \cite{radford2019language, roller-etal-2021-recipes, raffel2020exploring, rashkin-etal-2021-increasing} on dialogue generation. If trained on \newdata{}, we find that they are significantly more faithful while also enhancing other dialogue aspects like cooperativeness, creativity, and engagement.
These benefits also generalize to other knowledge-grounded datasets like CMU-DoG \cite{zhou-etal-2018-dataset} and TopicalChat \cite{Gopalakrishnan2019} in a zero-shot transfer setting.

\newdata also provides supervision for hallucination critics, which discriminate whether an utterance is faithful or not. We source positive examples from \newdata{} and negative examples from \noisydata{}. 
Compared to other dialogue inference datasets \citep{welleck2019dialogue,nie-etal-2021-like}, the classifiers trained on this data (which we call \critic) transfer better to general NLU tasks like MNLI \cite{williams-etal-2018-broad} and achieve state-of-the-art on BEGIN \cite{dziri2021evaluating}, a dialogue-specific knowledge grounding benchmark in a zero-shot setting.

Thus, \newdata{} holds promise to encourage faithfulness in information-seeking dialogue and make virtual assistants both more trustworthy. We release data and code for future research.\footnote{\href{https://mcgill-nlp.github.io/FaithDial/}{https://mcgill-nlp.github.io/FaithDial/}}

\section{\newdata: Dataset Design}
\label{sec4:annotation_pipeline}

Given the motivations adduced above, the primary goal of this work is to create a resource for faithful knowledge-grounded dialogue that allows for both training high-quality models and measuring the degree of hallucination of their responses. 
We define the notion of faithfulness formally as follows:

\begin{definition}[Faithfulness]
Given an utterance $u_{n}$, a dialogue history $\mathcal{H}=(u_1, \dots, u_{n-1})$, and knowledge $\mathcal{K}=(k_1, \dots, k_{j})$ at turn $n$, we say that $u_n$ is faithful with respect to $\mathcal{K}$ iff the following condition holds:
\begin{itemize}[noitemsep]
\item $\exists \, \Gamma_n$ such that $\Gamma_n \vDash u_n$, where $\vDash$ denotes semantic consequence and $\Gamma_n$ is a non-empty subset of $\mathcal{K}_n$. In other words, there is no interpretation $\cal I$ such that all members of $\Gamma_n$ are true and $u_n$ is false. 
\end{itemize}
\label{def:faithfulness}
\end{definition}
\noindent
Hence, an utterance can optionally be grounded on multiple facts but not none.

In what follows, we present the design of our task as well as our annotation pipeline to curate \newdata.
In our dialogue setting, we simulate interactions between two speakers: an information \seeker and a bot \expert. 

\begin{definition}[\textsc{Information Seeker}: A Human] \label{def:Information Seeker}
The information \seeker, a human, aims at learning about a specific topic in a conversational manner. They can express subjective information, bring up a new set of facts independent from the source $\mathcal{K}$, and even open up new sub-topics.
\end{definition}

From the perspective of Definition~\ref{def:Information Seeker}, utterances pronounced by the \seeker{} have a large degree of freedom. For example, the human can chat about personal life and can ask a diverse set of questions. On the other hand, the \expert{} is more restricted on what they can communicate.  %

\begin{definition}[\textsc{Wizard}: A Bot]

\label{Domain Expert}
The Wizard, a bot, aims at conversing in a knowledgeable manner about the \seeker's unique interests, 
 resorting exclusively to the available knowledge $\mathcal{K}$. They can reply to a direct question or provide information about the general topic of the conversation.\footnote{To encourage naturalness in the response, annotators were also asked to express empathy such as ``\textit{I'm sorry about ...''}. in case the \textsc{Seeker} expresses a very unfortunate event.}
\end{definition}

From Definition~\ref{Domain Expert}, it follows that there are three key rules the bot must abide by: first, it should be truthful by providing information that is attributable to the source $\mathcal{K}$. Second, it should provide information conversationally, i.e., use naturalistic phrasing of $\mathcal{K}$, support follow-up discussion with questions, and prompt user's opinions. Third, it should acknowledge its ignorance of the answer in those cases where $\mathcal{K}$ does not include it while still moving the conversation forward using $\mathcal{K}$.

\ignore{
Based on these definitions, the Seeker in  \noisydata{} is mapped to the \seeker \ (human) and the Wizard is mapped to the \expert \ (bot).
The main goal of \newdata\ benchmark is to simulate real-world conversations between an information-seeking human and a virtual assistant. Thus, we resolve the mismatch found in current dialogue crowdsourcing approaches, where collected human--human conversations serve as examples to model human--machine interactions.
}

\subsection{Data Selection}
Rather than creating a novel benchmark from scratch, however, we opt for fixing problematic utterances (which are the majority) in existing dialogue benchmarks \cite{dziri-etal-2022-origin}. The reason is three-fold: 1) while mostly hallucinated, existing datasets still contain useful faithful information.
2) as correction is faster than creation from scratch, this enables us to annotate examples on a larger scale; 3) two versions of the same dialogue turn, either hallucinated or faithful, can provide signal for (contrastive) learning and evidence for a linguistic analysis. In particular, we focus on \noisydata{} as our benchmark backbone.

 Initial pilot study revealed that \noisydata{} dialogues are more suitable for editing compared to other prominent knowledge-grounded dialogue benchmarks: TopicalChat \cite{Gopalakrishnan2019} and CMU-DoG \cite{zhou-etal-2018-dataset}.
In fact, according to \newcite{dziri-etal-2022-origin}, as shown in \Cref{tab:benchamrk_hall}, \noisydata{} is relatively less hallucinated compared to CMU-DoG and TopicalChat.
Moreover, full hallucinations---responses that contain no faithful content and that therefore need to be entirely thrown out--- are highly prevalent in the latter two ($61.4\%$ in CMU-DoG and $46.8\%$ in TopicalChat and only $19.7\%$ in \noisydata{}).
Moreover, knowledge snippets in \noisydata{} tend to be shorter, which is preferable as longer knowledge is correlated with increased hallucination due to the constrained cognitive capacity for text navigation and comprehension in humans \cite{de2010cognitive, destefano2007cognitive}.

\begin{table}[t]
\centering
\setlength{\tabcolsep}{2pt}
\scriptsize
\begin{tabular}{ c l | c c c    c  c  c  c c}
 \toprule
 & \multirow{2}{*}{\textbf{Dataset}} &\multirow{2}{*}{\textbf{Generic}} & \multicolumn{2}{c}{\textbf{Hallucination}}&\multicolumn{2}{c}{\textbf{Entailment}}  \\
  &&  & {{ \textbf{Full}}}  &  {{ \textbf{Partial}}} &  {{ \textbf{Faith.}}} &  {{ \textbf{Uncoop.}}} \\
  \midrule
 & \small \textbf{WoW}  & {\small \phantom  05.3} & {\small 19.7}& {\small 42.3}  & \small 24.1  & \small 8.5  \\ 
  &  {\small \textbf{CMU}} & {\small 13.2}  & {\small 61.4}  & {\small \phantom 05.1} & {\small 16.2}  & \small 4.1 \\
  & \multirow{1}{*}{\small \textbf{Topical}} &  \small 12.7 & \small 46.8 & \small 17.1 &  \small 22.9 & \small 0.5  
  \\
  \bottomrule
\end{tabular}
\caption{\small The breakdown of responses from \textsc{WoW}, CMU-DoG and TopicalChat according to BEGIN taxonomy \cite{dziri2021evaluating}. ``Faith." refers to faithful responses and ``Uncoop." refers to faithful but uncooperative responses given the conversation history. }
\label{tab:benchamrk_hall}
\end{table}%

 Our first step consists in filtering out \noisydata{} conversations where ground-truth knowledge $\cal K$ was not given, and annotators relied on personal knowledge instead. 
 Then, we focus on \seeker-initiated conversations and sample 44\% from the train (4094 conversations) and 100\% from validation (764 conversations) and 100\% from test (791 conversations).\footnote{We use the original \noisydata{} splits. Please note that only the training set in \newdata is smaller than \noisydata training set because of limited budget. The main goal of this paper is to provide a high-quality faithful dialogue benchmark rather than providing a large-scale dataset for training.}
 
\subsection{Crowd-sourced Annotations} Following the guidelines for ethical crowdsourcing outlined in \citet{sheehan2018crowdsourcing}, 
we hire Amazon Mechanical Turk (AMT) workers to edit utterances in \noisydata{} dialogues that were found to exhibit unfaithful responses.\footnote{To ensure clarity in the task definition, we provided turkers with detailed examples for our terminology. Moreover, we performed several staging rounds over the course of several months.
See the full set of instructions in Appendix \S\ref{huma_ann_amt}, the pay structure in Appendix \S\ref{appendix:pay},  and details about our quality control in Sec. \ref{data_quality} and Sec. \ref{sec_human_val}.}
First, workers were shown dialogues from \noisydata{} and asked to determine whether the \expert utterances are faithful to the source knowledge. 
To guide them in this decision, they were additionally requested to identify the speech acts (\vrm{} taxonomy; \citealt{stiles1992describing}) such as disclosure, edification, question, acknowledgment, etc; and the response attribution classes (\BEGIN{} taxonomy; \citealt{dziri2021evaluating}) such as hallucination and entailment for each of the  \expert's utterances according to \newcite{dziri-etal-2022-origin}'s schema.

\subsubsection{Editing the Wizard's Utterances}
Workers were instructed to edit the  \expert's utterances in the following cases, depending on their faithfulness.

\vspace{6pt}
\noindent
\textbf{Hallucination}.
They should remove information that is unsupported by the given knowledge snippet $\mathcal{K}$, and replace it with information that is supported. 
To ensure that the responses are creative, we disallowed workers from copying segments from $\mathcal{K}$. They were instead instructed to paraphrase the source knowledge as much as possible without changing its meaning  \cite{ladhak2021faithful, lux2020truth, goyal-durrett-2021-annotating}.  
If the inquiry of the \seeker cannot be satisfied by the knowledge $\mathcal{K}$, the \expert should acknowledge their ignorance and carry on the conversation by presenting the given knowledge in an engaging manner. In the example shown in Table~\ref{tab:trustdial-exmple}, the new \expert confirms that it cannot surf and instead enriches the conversation by talking about surfing as opposed to the original \expert{} who hallucinates personal information.

\vspace{6pt}
\noindent
\textbf{Generic} utterances such as ``\textit{That's nice}'' should be avoided solely on their own. Workers are instructed to enrich these responses with content that is grounded on the knowledge.

\vspace{6pt}
\noindent
\textbf{Uncooperativeness} If the response was determined to be faithful but uncooperative with respect to the user's requests, workers are required to make it coherent with the dialogue history while keeping it faithful.

\subsubsection{Editing the Seeker's Utterances}
Although the \seeker has no restrictions on their utterances, it is inevitable that the conversation may drift away--- because of the edits on the \expert's response---making the existing \seeker's next utterance in \noisydata{} incoherent with the new context.
In these cases, 
they perform edits on the \seeker's next utterance to make it coherent.
Consider Table~\ref{tab:trustdial-exmple} where workers had to edit the \noisydata{} \seeker's utterance as it was not coherent anymore with the freshly edited  \expert's response.

\section{Dataset Quality}
\subsection{Crowdworker Quality Control}
\label{data_quality}
To be eligible for the task, workers have to be located in the United States and Canada and have to answer successfully 20 questions as part of a qualification test.  Before launching the main annotation task,  we perform a small pilot round ($\sim$60 HITS) to check the performance of the workers. If we observe any errors, we email the concerned workers and provide them with examples on how to fix their mistakes in future HITS. Workers are also encouraged to reach out to us in case they find annotating a particular example ambiguous. At the end of the pilot round, we revoke access for workers who provide poor quality annotations.  
After several staging rounds, we launch the main annotation stage.
To ensure the quality does not drop, a linguistics major student evaluates the performance of workers daily (10 HITS on average per worker) and rejects poor quality work. Repeated mistakes result in the worker being blocked from the task entirely. In total, we ended up recruiting 10 well-trained workers. 
We also perform automatic quality control checks to enforce workers to avoid copying segments from the source knowledge. %
  
  \begin{table}[t]
\scriptsize
\centering
  \label{tab:docBetter}
\begin{tabular}{ l | l l l l l }
    \toprule
      {\textbf{Dataset}} &{\textbf{Train}} & {\textbf{Valid}}& {\textbf{Test}} \\
    \midrule
    Turns &  {\small 36809}  & {\small 6851}  &{\small 7101} \\
   Conversations &{\small 4094}  & {\small 764}  & {\small 791}  \\
 \scriptsize{Avg. Tokens for \textsc{Wizard}} & {\small 20.29}  & {\small 21.76}  & {\small 20.86}   \\
     Avg. Tokens for \textsc{Seeker} & {\small 17.25} & {\small 16.65}  & {\small 16.49}  \\
     Avg. Tokens for \textsc{Knowledge}& {\small 27.10}  & {\small 27.17}  & {\small 27.42} \\

     Turns per Conversation& {\small 9}  & {\small 9}  & {\small 9}    \\
    \bottomrule
  \end{tabular}
  \caption{Dataset statistics of \newdata.}
    \label{tab:statistics_data}
\end{table}

\begin{table*}[ht]
\footnotesize
    \centering
    \begin{tabular}{L{7cm} L{7cm}}
    \toprule
   \textbf{WoW} & \textbf{\newdata} \\
    \midrule
           \textbf{Seeker}: I would love to learn to surf but I'm terrified of sharks! Do you surf?
        & \textbf{Seeker}: I would love to learn to surf but I'm terrified of sharks! Do you surf? \\ \\
      \multicolumn{2}{L{15cm}}{\cellcolor{Gray} {\textbf{Knowledge}: Surfing is a surface water sport in which the wave rider, referred to as a surfer, rides on the forward or deep face of a moving wave, which is usually carrying the surfer towards the shore.}} \\ \\
       \textbf{Wizard}:    \textcolor{red}{I absolutely love to surf}, just riding on the forward face of a moving wave   \textcolor{red}{is so exhilarating}. Are there a lot of sharks near you?  
       \textcolor{violet}{(BEGIN: Partial hallucination, VRM: disclosure)}
        & \cellcolor{celadon}{\textbf{Wizard}: Unfortunately, it's impossible for me to surf. Remember, I'm just your virtual bot! Sorry to hear you're terrified from sharks. However, did you know that a surfer is someone who can ride on either the forward or deep face of a wave which typically sends her/him close to the shore?}\\ 
        \\
     \textbf{Seeker}: \textcolor{red}{I have no idea how many sharks are in the Gulf of Mexico but there out there! lol Where do like to surf?} \textcolor{teal}{(incoherent with the freshly edited Wizard response)}
        & \cellcolor{celadon}{\textbf{Seeker}:  Yeah, what scares me is the sharks, is there any non-ocean waves out there? }\\ \\
          \multicolumn{2}{L{15cm}}{\cellcolor{Gray} {\textbf{Knowledge}: Waves suitable for surfing are primarily found in the ocean, but can also be found in lakes or in rivers in the form of a standing wave or tidal bore.}} \\ \\
       \textbf{Wizard}: \textcolor{red}{There's a little hidden cove near me that has great waves, although waves suitable for surfing are primarily found in the Pacific ocean.} Have you ever watched people surf?   \textcolor{violet}{(\textsc{BEGIN}: hallucination, \textsc{VRM}: disclosure \& edification)}
       & \cellcolor{celadon}{\textbf{Wizard}: For sure there is. An alternative to ocean waves could be lakes and rivers which have a phenomenon called standing waves. Have you ever watched people surf?}\\ 
        \bottomrule
    \end{tabular}
    \caption{ \small A dialogue example showing the process of editing \noisydata \ utterances  to convert them to \newdata \ utterances. Text highlighted in red indicates hallucinated content. Text in violet indicates the BEGIN labels and the speech act VRM labels as identified by annotators. }
    \label{tab:trustdial-exmple}
\end{table*}
\subsection{Human validation} 
\label{sec_human_val}
To evaluate the quality of \newdata{}, we run two final rounds of annotations.
Firstly, we ask 3 new workers to edit the same 500 responses.  Since there is no straightforward way to measure inter-annotator agreement on edits, following \newcite{dziri-etal-2022-origin}, we measure the inter-annotator agreement on the identified response attribution classes (\BEGIN{}) and the speech acts (\vrm{}).
We report an inter-annotator agreement of 0.75 and 0.61 Fleiss' $\kappa$, respectively, which shows substantial agreement according to \citet{landis1977measurement}. This is an indicator of overall annotation quality: if the worker can reliably identify speech acts, they generally also produce reasonable edits.
Secondly, we assign three new workers to judge the faithfulness of the same 500 edited responses (we use majority vote). 
Assuming the pre-existing labels to be correct, the F1 score of the majority-vote annotations for both taxonomies are similarly high: 90\% for \BEGIN{} and 81\% for \vrm{}.  In total, we found that \newdata \ contains $94.4\%$ faithful responses and $5.6\%$ hallucinated responses, as shown in Figure \ref{fig:wizard_begin} (inner circle), and this shows the high quality of \newdata.

\section{Dataset Analysis}
\subsection{Dataset Statistics}
Overall, \newdata \ contains a total of 5,649 dialogues consisting of 50,761 utterances. Table \ref{tab:statistics_data} reports statistics for each dataset split. %
To curate \newdata, workers edited $84.7\%$ of the  \expert responses (21,447 utterances) and $28.1\%$ of the \seeker responses (7,172 utterances). 
In particular, 3.8  \expert turns per conversation were modified on average, as opposed to only 1.2 \seeker turns.
The low percentage of the \seeker edits shows that our method does not disrupt the cohesiveness of the conversations.

\begin{figure*}[t]
    \centering
     \subfigure[\label{fig:wizard_begin}\footnotesize \newdata]{\includegraphics[width=0.37
     \linewidth]{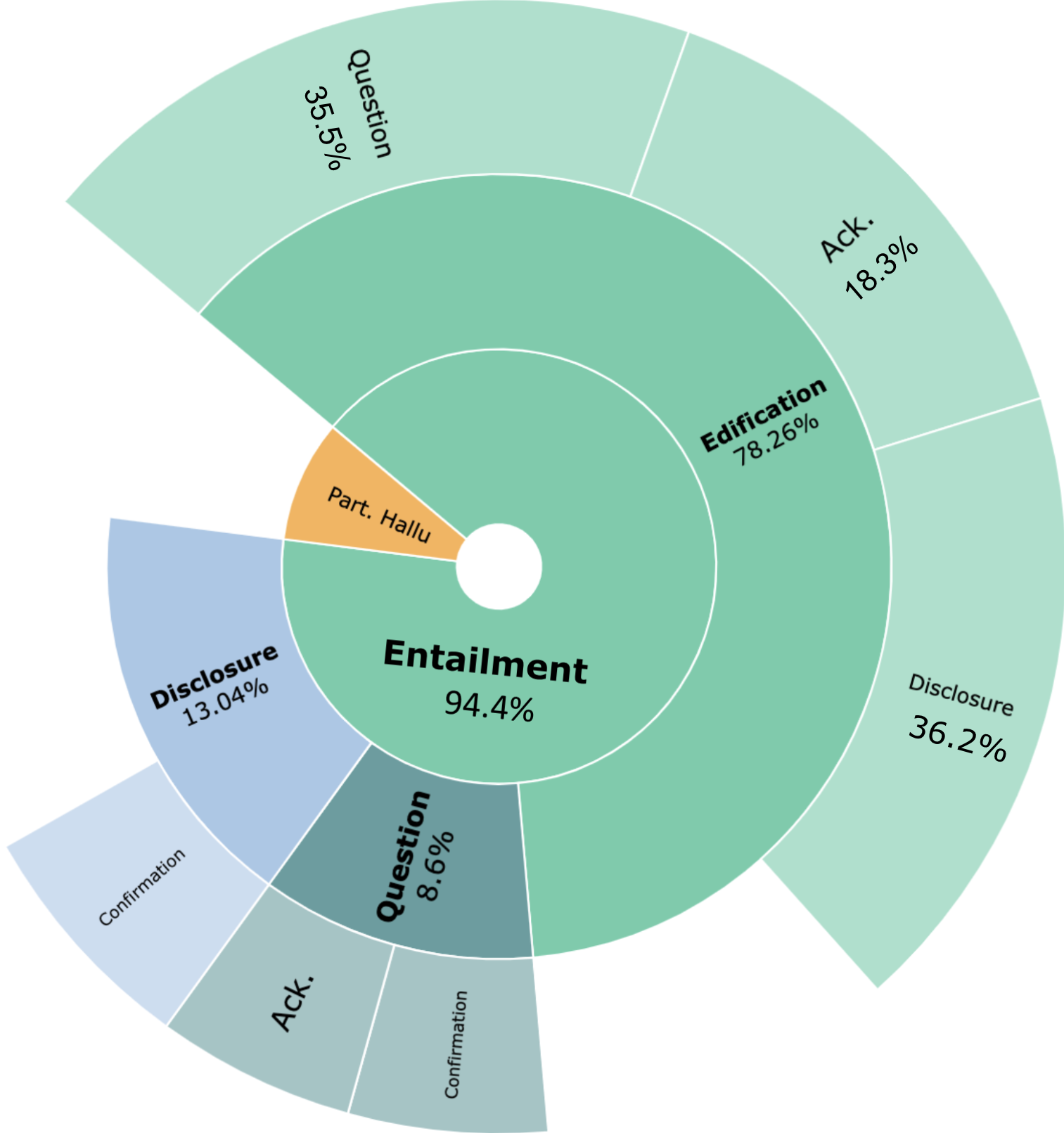}}
    \subfigure[\label{fig:cmu_begin}\footnotesize \noisydata]{\includegraphics[width=0.42\linewidth]{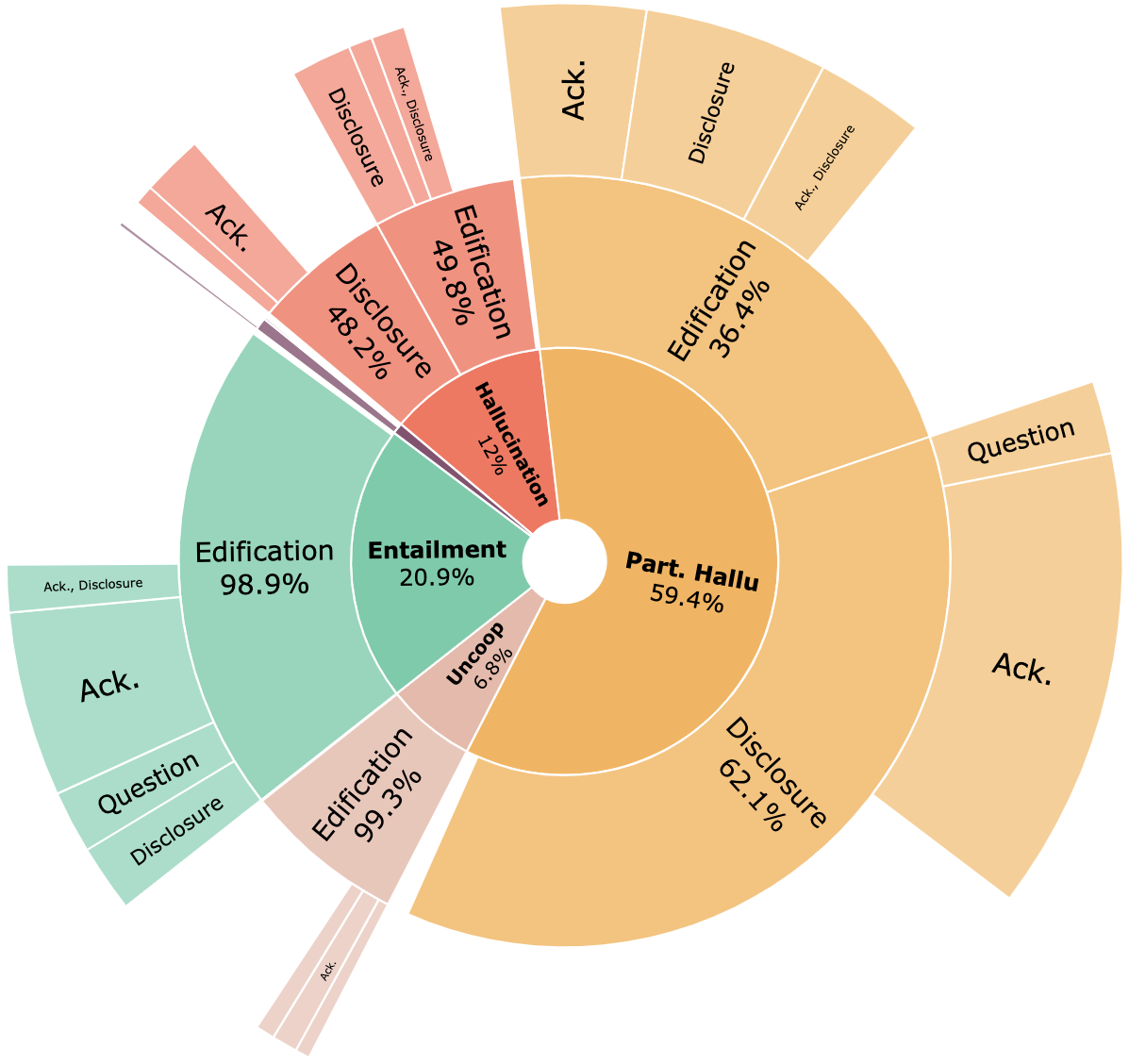}}
    \caption{\small Coarse-grained (\BEGIN) and fine-grained speech act (\vrm) \ distributions used by wizards in \newdata\ and \noisydata. The inner most circle shows the breakdown of coarse-grained types: Hallucination (red), Entailment (green), Partial Hallucination (yellow), Generic (purple), and Uncooperative (pink).  
    The outer circles show the fine-grained types of each coarse-grained type.}
    \label{fig:faithful_trustdial}
    \vspace{-8pt}
\end{figure*}
\subsection{Linguistic Phenomena}
\label{sec:linguistic}

\begin{figure}[t]
\centering
\includegraphics[width=\linewidth]{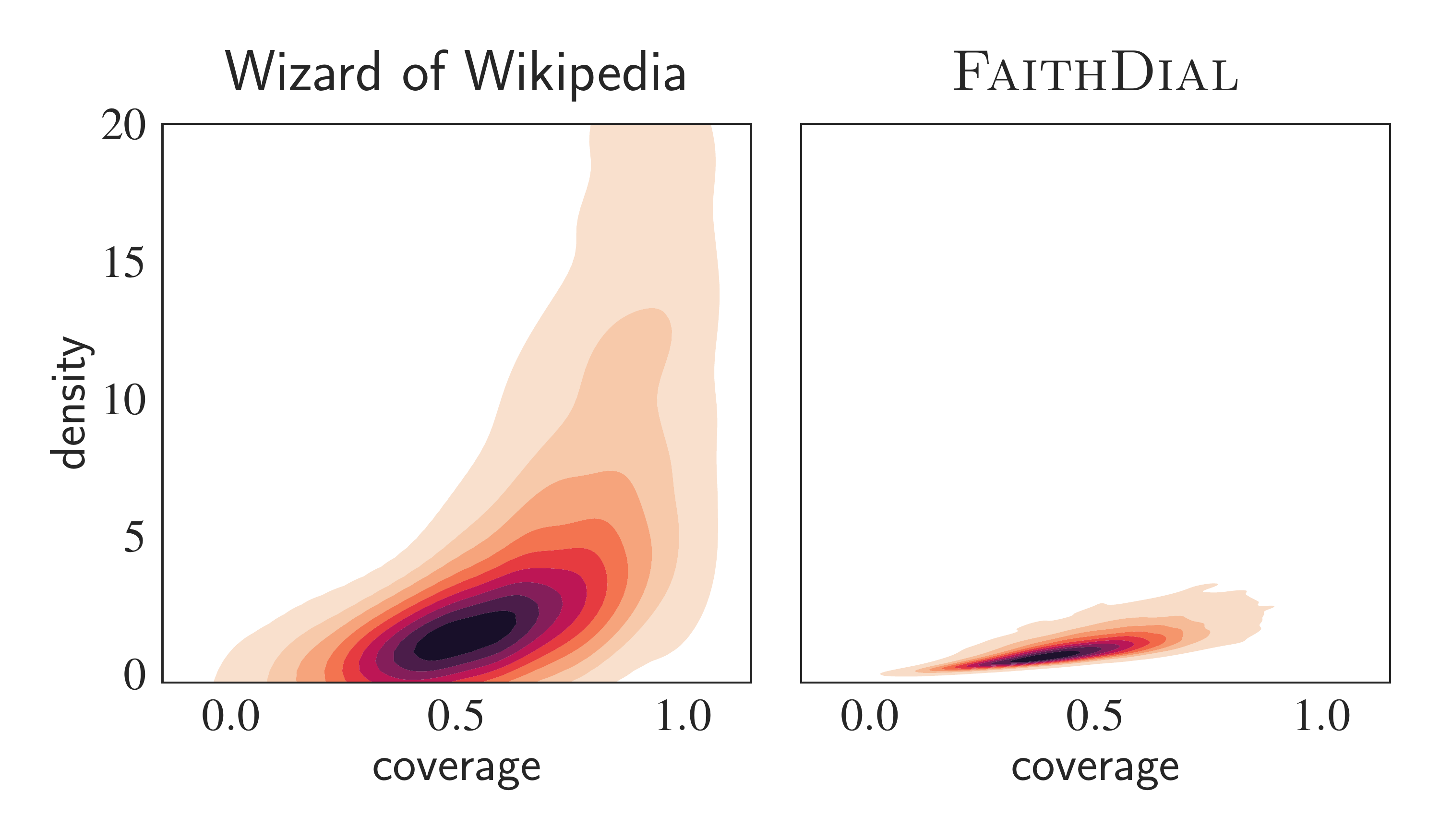}
\caption{Density and coverage in \noisydata{} \cite{dinan2018wizard} (left) vs.\ \newdata{} (right). Responses in \newdata\ tend to be abstractive to a large degree compared to \noisydata{}.}
\label{fig:abstractive}
\end{figure}

\subsubsection{Faithfulness}
Based on our human validation round of 500 examples, \newdata \ contains $94.4\%$ faithful responses and $5.6\%$ hallucinated responses.
On the other hand, our large-scale audit of the entirety of \noisydata reveals that it is interspersed with hallucination ($71.4\%$), with only a few faithful turns ($20.9\%$), as shown in Figure \ref{fig:cmu_begin} (inner circle). 
This finding is consistent with the analysis of \citet{dziri-etal-2022-origin} on a smaller sample. 
In our work, \newdata{} cleanses dialogues from hallucination almost entirely.

We also report the speech acts employed to ensure faithfulness in \newdata in the outer circle in Figure \ref{fig:faithful_trustdial}. We observe that  \expert resorts to a diverse set of speech acts to convey faithful information in a conversational style (see the Entailment pie): $78.26\%$ of the responses contain objective content (\textit{Edification}) that is interleaved with dialogue acts such as acknowledging receipt of previous utterance ($18.3\%$), asking follow-up questions ($35.5\%$), and sparking follow-on discussions by expressing opinions still attributable to the knowledge source ($36.2\%$). Moreover, the \expert used some of these very techniques, such as \textit{Disclosure} ($13.04\%$) and \textit{Questions} ($8.6\%$), in isolation. 
On the other hand, faithfulness strategies (see Entailment) in \noisydata \ are mostly limited to edification ($98.9\%$), curbing the naturalness of responses.

\subsubsection{Abstractiveness}
After establishing the faithfulness of \newdata{}, we investigate whether it stems from an increased level of extractiveness or abstractiveness with respect to the knowledge source.
Extractive responses reuse the same phrases as the knowledge source, while abstractive responses express the same meaning with different means.
Although extractive responses are an easy shortcut to achieving more faithfulness, it comes at the cost of creativity. Ideally, we want responses that are faithful as well as creative, meaning responses that are not just a copy paste of the knowledge but rather a creative use of it. 
To measure creativity, we borrow two metrics from \citet{grusky2018newsroom} designed to quantify the extractive and abstractive nature of summaries: \textit{Density} and \textit{Coverage}. Density represents the average length of the text spans copied from the knowledge that are contained in the response. Coverage instead measures the percentage of words existing in a response that are also found in the source knowledge. Figure \ref{fig:abstractive} illustrates the density and coverage distributions in \newdata (right) \ vs.\ \noisydata (left). 
We observe that while the coverage (x-axis) is similar in both \newdata{} and \noisydata{}, the density (y-axis) is always low in \newdata{} but often high in \noisydata{}.
This indicates that responses in \newdata\ tend to be abstractive to a large degree.%

Based on this, we also study which specific abstractive strategies \expert{} adopts to present knowledge from $\mathcal{K}$ without repeating long fragments. The strategies we discovered, fall into five broad categories: inference of new knowledge from $\mathcal{K}$, rewording, reshaping the syntactic structure, abridging long expressions, and introducing connectives. We discuss these categories in more detail in \S\ref{app:abstraction}.

\subsubsection{Fallback Responses in \newdata{}}
We further probe the \expert \ responses with respect to their  ability to handle unanswerable questions. %
We randomly sample 45 dialogues containing 400 responses and ask a linguist to annotate them. Overall, we found that $48\%$ of the conversations contain unanswerable utterances: 
On average $33\%$ of the \expert{} responses within the same conversation were edited to provide fallback responses. Out of those fallback responses, $30\%$ were triggered by personal questions, $50\%$ by objective questions about the topic, and $20\%$ by opinions. In these cases, to avoid interrupting the flow of the conversation, the \expert \ informs the \seeker\ about facts from the source knowledge besides acknowledging its ignorance of the right answer.

\section{Experiments}
The purpose of {\newdata} is two-fold: first, the collected labels can serve as training data for a critic to determine whether a given response is faithful or hallucinated.
The second goal is providing high-quality data to generate faithful responses in information-seeking dialogue. 
Given knowledge $\mathcal{K}_n$ and the conversation history $\mathcal{H}=(u_1, \dots, u_{n-1})$, the task is to generate a response $u_n$ faithful to $\mathcal{K}_n$. 
We benchmark a series of state-of-the-art dialogue models \cite{radford2019language, roller-etal-2021-recipes, raffel2020exploring, rashkin-etal-2021-increasing} on {\newdata}.
We also evaluate them on \noisydata{} and in a zero-shot transfer setup on CMU-DoG, and TopicalChat). We implement all the baselines using the Huggingface Transformers library \cite{wolf-etal-2020-transformers}.

\subsection{Task I: Hallucination Critic}
\label{critic}

We frame the problem of identifying hallucination as a binary classification task where the goal is to predict whether an utterance is faithful or not, given the source knowledge. This characterization of the problem is reminiscent of previous work \cite{dziri-etal-2019-evaluating, welleck-etal-2019-dialogue, nie-etal-2021-like} on detecting contradiction within a conversation.
 
For this purpose, we curate a dataset, \critic{}, derived from human annotations in \newdata. Specifically, we take 14k  \expert utterances from {\noisydata} labelled as hallucination (Section~\ref{sec4:annotation_pipeline}) as negative examples. The \expert responses from {\noisydata} labelled as entailment along with newly edited {\expert} utterances (20k in total) count as positive examples.
Overall, \critic{} consists of 34k examples for training. We compare the performance of models trained on \critic{} against models trained on two dialogue inference datasets ---DNLI \cite{welleck-etal-2019-dialogue} and DECODE \cite{nie-etal-2021-like}---and on a well-known natural language inference (NLI) dataset, MNLI \cite{williams-etal-2018-broad}. 
For all datasets, we choose RoBERTa$_\text{Large}$ \cite{liu2019roberta} as a pre-trained model. Implementation details can be found in \S\ref{sec:implement}.
We measure the transfer performance of different critics on MNLI, \BEGIN and \critic in zero-shot settings wherever possible.

The results are presented in Table~\ref{tab:critic}. 
In the zero-shot setting, the critic trained on \critic{} substantially outperforms the baselines on MNLI and BEGIN by a large margin, indicating that \newdata allows transfer to both a generic language understanding task as well as dialogue-specific knowledge grounding benchmark. On the other hand, the transfer performance of DECODE and DNLI are poor on both generic and dialogue-specific classification tasks.  
Surprisingly, MNLI transfers well to \critic.

\begin{table}[t]
\centering
\small
\begin{tabular}{ l | c  c  c}
 \toprule
 \multirow{2}{*}{\scriptsize{Trained on}} & \multicolumn{3}{c}{\footnotesize {Tested on}} \\
 & \bf \scriptsize MNLI &  \bf \scriptsize BEGIN & \bf \scriptsize \critic \\
 \midrule
    {\small \scriptsize DECODE} & {\small 62.5$^\dag$}  & {\small 58.8$^\dag$}  & {\small 38.5$^\dag$}   \\ 
     {\small \scriptsize DNLI} & {\small 52.4$^\dag$}   & {\small 59.8$^\dag$} & {\small 30.9$^\dag$} \\ 
     {\small \scriptsize MNLI}  & \bf 93.1 &  {\small 61.1$^\dag$} &  {\small 81.6$^\dag$} \\
\midrule
   {\small \scriptsize \critic}  & 74.7$^\dag$  & \bf {\small 71.6$^\dag$} & \bf {\small 86.5} \\
 \bottomrule
\end{tabular}
\caption{{\small Transfer results (accuracy) of the hallucination critics trained and tested on different datasets. $\dag$ indicates zero-shot transfer results.
}}
\label{tab:critic}
\end{table}

\ignore{
\begin{table*}[t]
\centering
\small
\begin{tabular}{ l | c c | c c | c c}
 \toprule
  & \multicolumn{2}{c|}{\bf MNLI} & \multicolumn{2}{c|}{\bf \critic} & \multicolumn{2}{c}{\bf BEGIN} \\
  & {{ \textbf{Acc}}} & \textbf{F1} & {{ \textbf{Acc}}} & \textbf{F1} & {{ \textbf{Acc}}} & \textbf{F1}   \\
 \midrule
 
  {\small \critic}  & 74.7 & 73.9 & {\small 86.5} & {\small 83.6} & {\small 71.6} & {\small 71.5}   \\
    \hline

     {\small DECODE} & {\small 62.5} & {\small 57.9} & {\small 38.5} & {\small 38.5}  & {\small 58.8} & {\small 55.5} \\ 
       
     \hline
     {\small DNLI} & {\small 52.4} & {\small 52.4}  & {\small 30.9} & {\small 27.1} & {\small 59.8} & {\small 51.5}  \\ 
     \hline
 {\small MNLI}  & 93.1 & 92.4 & {\small 81.6} & {\small 76.7} & {\small 61.1} & {\small 61.0}   \\

 \bottomrule
\end{tabular}
\caption{{\small Accuracy and macro F$_1$ measure of the hallucination critic, trained on various inference datasets, and evaluated zero-shot on {\BEGIN} \cite{dziri2021evaluating} and the test data from \critic.
}}
\label{tab:critic}
\end{table*}
}

\subsection{Task II: Dialogue Generation}
\begin{table*}[t]
\centering
\scriptsize
\renewcommand{\arraystretch}{0.9}
\begin{tabular}{ c l | c  l  l  c  c  c  c  c}
 \toprule
 & \multirow{2}{*}{\footnotesize \textbf{Models}} &\multirow{2}{*}{\footnotesize{\textbf{Critic $\downarrow$}}} & \multicolumn{2}{c}{\footnotesize \textbf{Q$^2$} $\uparrow$} &{\footnotesize \textbf{BERTScore}$\uparrow$}
 &{\footnotesize \textbf{F1}$\uparrow$}
 &{\footnotesize \textbf{BLEU}$\uparrow$}
 &{\footnotesize \textbf{ROUGE}$\uparrow$}\\
  &&  & {{\footnotesize \textbf{F1}}} &  {{\footnotesize \textbf{NLI}}} &
  {{ $(u, \mathcal{K})$}} & {{ $(u, \mathcal{K})$}} & {{ $(u, g)$}}&  {{ $(u, g)$}} \\
  \midrule
  \multirow{6}{*}{\rotatebox[origin=c]{90}{ \texttt{\textbf{WoW}}}} & {{\textsc{GPT2}}}  & {60.1}  & {42.2} & {51.4}& {0.29} & {47.7}  & {7.3} & {18.3} \\ 
    & \multirow{1}{*}{\textsc{DialoGPT}} & {59.4}  & {41.4}  & {52.5}   & {0.34}   & {53.5}&  {8.3} &  {29.5} \\
      &  {\textsc{DoHA}} & {53.2}  &  {63.3} &  {70.1}  & {0.32} & {56.1}&   {9.4}&  {32.3}
  
  \\
  &  {{\textsc{T5}}} & {{46.5}}  &  {67.7}  &  {75.2} & {0.41}  &  {61.7} &  {9.5} &  {32.9} \\

   \cline{2-9}
    & \cellcolor{Gray}{ \textsc{T5-CTRL}} & \cellcolor{Gray}{45.2}  &  \cellcolor{Gray}{70.3} &  \cellcolor{Gray}{76.2}  & \cellcolor{Gray}{\textbf{0.45}} & \cellcolor{Gray}{\bf65.2}
    &\cellcolor{Gray}{\textbf{9.9}} & \cellcolor{Gray}{33.1}
  \\ 
 
       &  \cellcolor{Gray}{  \textsc{T5-LossTruncation}} & \cellcolor{Gray}{\textbf{41.4}}  &  \cellcolor{Gray}{\textbf{71.2}} &  \cellcolor{Gray}{\textbf{79.4}}  & \cellcolor{Gray}{0.43} &\cellcolor{Gray}{65.0}
       &\cellcolor{Gray}{9.8} & \cellcolor{Gray}{\textbf{33.4}}
  \\ 
  \midrule
 \multirow{7}{*}{\rotatebox[origin=c]{90}{ \texttt{\textbf{FaithDial}}}} & {{\textsc{GPT2}}}  & 5.8 & 58.4 & {69.8} & {0.36} & {50.4}& {9.5}&   33.4   \\ 
   & \multirow{1}{*}{\textsc{DialoGPT}} &      {5.6} &  {56.5}  &  {66.2}  &  {0.36}  &  {52.3}   &  {9.6}  &  {33.1} 
 
   \\ 
    & \multirow{1}{*}{\textsc{DoHA}} &  {4.9} &  {69.1} &  {78.3}  & {0.39}  &  {58.3} &  {9.9} &  31.8 
  \\

  &  {{\textsc{T5}}} & {4.3} & {70.4}  &  {79.5}&{0.41} &  {59.2} &  {10.3} &  {33.9} \\
    \cline{2-9}
      & \cellcolor{Gray}\multirow{1}{*}{ \textsc{T5-CTRL}} &\cellcolor{Gray}{5.7} &  \cellcolor{Gray}{\bf72.4} & \cellcolor{Gray}{\textbf{81.5}}  &\cellcolor{Gray}{\textbf{0.46}}  & \cellcolor{Gray}{\textbf{62.2}} &\cellcolor{Gray}{10.4} & \cellcolor{Gray}{33.9}  \\ 
        &  \cellcolor{Gray}{  \cellcolor{Gray}\textsc{T5-LossTruncation}} & \cellcolor{Gray}{4.0}  &\cellcolor{Gray}{71.9} &\cellcolor{Gray}{80.2}  &\cellcolor{Gray}{0.42} & \cellcolor{Gray}{59.1}& \cellcolor{Gray}{10.2}& \cellcolor{Gray}{33.9} 
  \\ 
  &  \cellcolor{Gray}{{ \textsc{T5-InfoNCE}}} & \cellcolor{Gray}{\textbf{1.4}} & \cellcolor{Gray}{{70.8}}  & \cellcolor{Gray}{{80.9}}  & \cellcolor{Gray}{0.39} & \cellcolor{Gray}{55.8} & \cellcolor{Gray}{\textbf{10.9}} & \cellcolor{Gray}{\textbf{35.8}} \\
    \midrule
    \multirow{7}{*}{\rotatebox[origin=c]{90}{{\begin{tabular}[c]{@{}c@{}}\texttt{\textbf{\footnotesize FaithDial}}\\ \texttt{\textbf{\scriptsize (+WoW)}}\end{tabular}} }} & {{GPT2}}  & 7.2 &  62.3 & {73.4} & {0.39} & {54.2}& {10.0}&     34.2 \\ 
   & \multirow{1}{*}{\textsc{DialoGPT}} &  8.2   &   54.5 & 65.6   &   0.42 &  48.6   &  {\phantom 08.9}  &  32.3
  \\ 
     & \multirow{1}{*}{DoHA} & {1.6}&  {66.7} &  {77.4}  & {0.40}  & 55.8  & 11.4 & 36.5
  \\ 
  &  {{T5}} & 2.0 & {70.2} & {80.1}  &  {0.41} & {57.5} &  {11.5}  & \textbf{37.2} \\

    \cline{2-9}
      & \cellcolor{Gray}\multirow{1}{*}{T5-CTRL} & \cellcolor{Gray}{4.5}&  \cellcolor{Gray}{\textbf{73.4}} &  \cellcolor{Gray}{\textbf{83.5}}  &\cellcolor{Gray}{\textbf{0.50}}  & \cellcolor{Gray}\textbf{64.6} & \cellcolor{Gray}{10.9}&\cellcolor{Gray}{35.6} 
  \\ 

   & \cellcolor{Gray}{\textsc{T5-LossTruncation}} & \cellcolor{Gray}{4.0}  &  \cellcolor{Gray}{70.2} &  \cellcolor{Gray}{79.1}  & \cellcolor{Gray}{0.41} & \cellcolor{Gray}{58.9}&\cellcolor{Gray}{10.4}& \cellcolor{Gray}{33.9} 
  \\
  &  \cellcolor{Gray}{T5-InfoNCE}  & \cellcolor{Gray}{\textbf{1.4}}  & \cellcolor{Gray}{69.8} &  \cellcolor{Gray}{79.8}  & \cellcolor{Gray}{0.40} & \cellcolor{Gray}{57.1}& \cellcolor{Gray}{\textbf{11.5}}& \cellcolor{Gray}{36.5}\\

  \bottomrule
\end{tabular}
\caption{\small Model performance on the test split of \newdata{}. Metrics measure either the degree of hallucination of generated responses $u$ with respect to knowledge $\cal K$ or their overlap with gold faithful responses $g$. Gray blocks correspond to models that are specifically designed to alleviate hallucinations. Note that we do not use InfoNCE for models trained on \noisydata{} as positive examples are not available in this setting. %
}
\label{tab:metrics_eval}
\vspace{-2mm}
\end{table*}

\subsubsection{Methods}
For the task of dialogue generation, we consider a series of state-of-the-art models ranging from general-purpose LMs---such as GPT2 \cite{radford2019language}, \textsc{DialoGPT} \cite{zhang-etal-2020-dialogpt}, and T5 \cite{raffel2020exploring}---to models that are  specifically designed to provide better grounding, such as DoHA \cite{prabhumoye-etal-2021-focused}, or to alleviate hallucination, such as CTRL \cite{rashkin-etal-2021-increasing}.  
DoHA augments BART \cite{lewis-etal-2020-bart} with a two-view attention mechanism that separately handles the knowledge document and the dialogue history during generation. CTRL equips LMs with control tokens ($<$\texttt{\small 
\small objective-voice}$>$, $<$\texttt{\small lexical-overlap}$>$, and $<$\texttt{\small entailment}$>$) whose embeddings are learned at training time. At test time, these steer a model towards generating utterances faithful to a source of knowledge. Finally, we adopt a training strategy, called loss truncation \cite{kang-hashimoto-2020-improved} to cope with the presence of hallucination in \noisydata{}, by adaptively eliminating examples with a high training loss.

In addition to existing models, we also consider an auxiliary objective to attenuate hallucination during training \cite{cao-wang-2021-cliff, tang2021confit}. In particular, we adopt InfoNCE \cite{oord2018representation}, a contrastive learning loss, to endow models with the capability of distinguishing faithful responses $\mathbf{x}^+$ from hallucinated ones $\mathbf{x}^-$. Given an embedding of the context $\mathbf{c}$, which includes both conversation history and knowledge:
\begin{equation}
    \mathcal{L}_{\mathrm{InfoNCE}} = - \log \frac{\exp (\mathbf{c}^\top \mathbf{x}^+)}{\sum_{\mathbf{x}^\prime} \exp (\mathbf{c}^\top \mathbf{x}^\prime)}
\end{equation}
To generate up to $k=8$ negative candidates $\mathbf{x}^-$, we follow a perturb-and-generate strategy for each utterance in the training data. More precisely, we manipulate the gold knowledge snippets to alter their meaning and feed them along with the history to an auto-regressive model fine-tuned on \noisydata{}. We use two perturbation techniques proposed by \citet{dziri2021evaluating}: verb substitution and entity substitution. Additionally, utterances labelled as hallucination by human annotators in \noisydata{} are also included in the negative samples. The implementation details and hyperparameters are provided in \S\ref{sec:implement}.

\begin{table*}[ht]
\footnotesize
    \centering
    \begin{tabular}{L{7cm} L{6.8cm}}
    \toprule
  \textbf{Context} & \textbf{Responses} \\
    \midrule
  
        \multirow{6}{7cm}{\textbf{Knowledge}: A paramedic is a healthcare professional, predominantly in the pre-hospital and out-of-hospital environment, and working mainly as part of emergency medical services (EMS), such as on an ambulance.
        \\ \textbf{History}: \colorbox{gray}{Do you like medical science or have you}  \colorbox{gray}{considered being a paramedic?}} & \textbf{\textsc{T5-LossTruncation-WoW}}: \textcolor{red}{I'm a nurse and I know a lot about EMS. I'm in the hospital and it can be tiring}. (\underline{Hallucination}) \\
        & \textbf{\textsc{T5-InfoNCE-FaithDial}}:  I am a bot so I can't be a paramedic. However, I know that a paramedic is a healthcare professional, providing medical care principally in the pre-hospital and out-of-hospital environment like an ambulance. (\underline{Faithful}) \\

        \midrule

        \multirow{6}{7cm}{\textbf{Knowledge}: In the Middle Ages and Renaissance, green was the color commonly associated with merchants, bankers and the gentry, while red was reserved for the nobility.
        \\ \textbf{History}: \colorbox{gray}{Yes, that is camouflage.  It is the color most closely} \colorbox{gray}{associated with spring, nature, health, youth, hope, } \colorbox{gray}{and envy. I love green. It is so versatile.}}
        &\textbf{\textsc{T5-CTRL-WoW}}: \textcolor{red}{Green is so nice to have. It brings joy and excitement.} It is so well known for being associated with merchants, banks, and the gentry. (\underline{Partial Hallucination}) 
        \\
        &
        \textbf{\textsc{T5-CTRL-FaithDial}}: Oh, cool!  Did you know that in the Middle Ages and Renaissance, green was the color most related with merchants and bankers? (\underline{Faithful})\\ \\
        \bottomrule
    \end{tabular}
    \caption{\small Sample responses from different models.  Models trained on \newdata \ have a higher success rate in providing faithful responses as opposed to the ones trained on \noisydata. Text highlighted in red indicates hallucination.}
    \label{tab:unanswerable_responses}
    \vspace{-3mm}
\end{table*}
\subsubsection{Automatic Evaluation}
We rely on several metrics that provide a multi-faceted measure of performance. A first group measures the degree of hallucination of generated responses.  
The \textbf{Critic} model trained on \critic (Section~\ref{critic}) returns the percentage of utterances identified as unfaithful. %
\textbf{Q$^2$} \cite{honovich-etal-2021-q2} measures faithfulness via question answering.
It takes a candidate response as input and then generates  corresponding questions. Then, it identifies possible spans in the knowledge source and the candidate response to justify the question--answer pairs \cite{durmus-etal-2020-feqa, wang2020asking}. 
Finally, it compares the candidate answers with the gold answers, in terms of either token-level \textbf{F1} score or a \textbf{NLI}-inspired similarity score based on a RoBERTa model.
\textbf{BERTScore} \cite{zhang2019bertscore} rates the \textit{semantic} similarity between the generated response $r$ and the knowledge $\mathcal{K}$ based on the cosine of their sentence embeddings. \textbf{F1} measures instead the token-level \textit{lexical} overlap between $u$ and $\mathcal{K}$. 
Finally, as a second set of metrics, we report BLEU \cite{papineni2002bleu} and ROUGE \cite{lin-2004-rouge}, which reflect instead the n-gram overlap between $u$ and the gold (faithful) response $g$.

\paragraph{WoW vs \newdata.}
In order to evaluate the ability of {\newdata} to reduce hallucination in generated responses,
Table \ref{tab:metrics_eval} illustrates three experimental setups with different training data. \noisydata{} corresponds to the first block and {\newdata} to the second block. The third block reflects a hybrid setup where a model is fine-tuned sequentially on {\noisydata} as an intermediate task and then on {\newdata}. 
We evaluate all on the {\newdata} test set.

We find that training on {\newdata} yields a substantial reduction in hallucination. For example, \textsc{T5} trained on \newdata \ decreases hallucination by $42.2$\% according to the Critic and increases the faithfulness score (Q$^2$-NLI) by $4.3$\% compared to \textsc{T5} trained on \noisydata.\footnote{The relatively high score of \textsc{T5-WoW} on Q$^2$-NLI may be due to this metric not being robust to \textit{partial} hallucinations.} This corroborates the prominence of data quality compared to the data quantity ({\newdata} is one third of \noisydata).
When initializing the models trained on {\newdata} with the noisy checkpoint from {\noisydata} (third block), we observe a performance boost in all models across all metrics, except a marginal drop in Critic for \textsc{GPT2} and \textsc{DialoGPT}. This shows that models can extract some useful conversational skills from \noisydata{} despite its noisy nature.

\begin{table*}[t]
\scriptsize
\centering
\begin{tabular}{ c l | c  |c | l  l  l | l }
 \toprule
  &\multirow{2}{*}{\scriptsize \textbf{Models}} & \multirow{2}{*}{\scriptsize{\textbf{Interpretable}}} 
   & \multirow{2}{*}{\scriptsize\textbf{Hallucination}}
  & \multicolumn{3}{c|}{\scriptsize\textbf{Faithfulness}} 
 & \multirow{2}{*}{\scriptsize \textbf{Generic}}
 \\
  &&&  &
   {{\scriptsize \textbf{Coop.}}}  & {{  {{\scriptsize \textbf{Abst.}}} }} &  {{  {{\scriptsize \textbf{Enga.}}} }}\\
  \midrule

 \multirow{3}{*}{\rotatebox[origin=c]{90}{ \texttt{\scriptsize\textbf{WoW}}}} &{{\textsc{T5}}}  & 93.2\%  & {{\phantom 055.8\%}$^{**}$} & {2.97$^{*}$} & {1.95$^{*}$}  & 1.72$^{*}$ &  {2.2\%}  \\ 
  &{{\textsc{T5-CTRL}}}  & {95.2\%}  & {{44.2\%}$^{*}$} & {1.97$^{*}$} & {0.92$^{*}$}  & {1.33$^{*}$} & {\textbf{0.9\%}}  \\ 
  &{{\textsc{T5-LossTruncation}}}  & {94.3\%}  & {{\phantom 042.5\%}$^{**}$} &  {2.87$^{*}$} & {1.87$^{*}$}  & 1.83$^{*}$ & {1.2\%}  \\ 
  \midrule
  \multirow{5}{*}{\rotatebox[origin=c]{90}{ \texttt{\scriptsize\textbf{FaithDial}}}}  & \multirow{1}{*}{\textsc{T5}} & {94.4\%}  & {23.2\%}$^{*}$   & 3.63   & {2.43$^{*}$}&  2.33 & 1.4\%
  \\
  
       &{{\textsc{T5-WoW}}} & {{95.2\%}}  & 20.9\%$^{*}$    & {3.59}  & 2.44 &  2.37 &  1.0\% \\
    &{{\textsc{T5-CTRL}}} & {{96.7\%}}  &  {20.8\%}$^{*}$   & {2.55$^{*}$}  &  1.42$^{*}$ &  2.10$^{*}$ &  1.0\% \\
    
     &{{\textsc{T5-LossTruncation}}} & {{94.2\%}}  &  24.2\%$^{*}$    & {3.59}  & 2.42$^{*}$ &  2.03$^{*}$ &  \bf{0.9\%} \\
      & {{\textsc{T5-InfoNCE}}} & {{\textbf{97.2\%}}}  & \bf 19.9\%  & {\bf 3.79}  & \bf 2.92 & \bf 2.60 & \bf 0.9\% \\

  \bottomrule
\end{tabular}
\caption{\small Human Evaluation on 1600 generated \newdata{} responses (200 $\times$ 8) from different models on  the  test data. $^{*}$ and $^{**}$ indicates that the results are significantly different from the best result in that column (bolded) with p-value $<$ 0.05, $<$ 0.01 respectively. `Coop.', `Abst.', and `Enga.' means cooperativeness, abstractiveness, and engagingness respectively.}
\label{tab:human_eval}
\end{table*}

\paragraph{Models.}
First, we observe that \textsc{T5} consistently performs favourably in reducing hallucination in all setups and across all metrics, compared to the rest of the vanilla baselines: \textsc{GPT2}, \textsc{DialoGPT}, and \textsc{DoHA}. 
Additionally, we compare models that are designed specifically to alleviate hallucination. Results are reported in the grey blocks of Table~\ref{tab:metrics_eval}.  We choose the best vanilla model \textsc{T5} as the backbone for \textsc{CTRL}, \textsc{InfoNCE} and \textsc{LossTruncation}. By virtue of these methods, faithfulness increases even further, which demonstrates their effectiveness. Sample responses from different models are presented in Table~\ref{tab:unanswerable_responses}.

\paragraph{Abstractiveness.}
We find that while \newdata{}, especially in the hybrid setup, increases the semantic similarity between generated responses and knowledge (BERTScore) by 7\% compared to \noisydata{}, the word overlap (F1) between them is almost unaffected. This indicates that \noisydata{} induces extractiveness over abstractiveness in models, which is not desirable. This is especially true for \textsc{T5-CTRL} variants, as their training objective encourages word overlap. Instead, we observe that \textsc{T5-InfoNCE} achieves both faithfulness and abstractiveness as it yields the lowest scores for hallucination ($1.4$ Critic) and extractiveness ($55.8$ F1).

\subsubsection{Human Evaluation}

In addition to the automated metrics, we conduct human evaluation to assess the presence of hallucination in models trained on \newdata, as well as other aspects in generated dialogues such as cooperativeness, engagingness, and abstractiveness.
Following \newcite{rashkin2021measuring}, our evaluation consists of a two-stage annotation process. 
First, the annotators are asked to determine whether responses are stand-alone (i.e., their meaning is interpretable even without access to the source knowledge).
If not, they are deemed to be too vague or ill-formed to judge their faithfulness. 
Second, if the response is interpretable, the annotators are requested to evaluate whether the response is grounded on the source knowledge.
If the response was deemed not faithful, we further ask the annotators to mark it as hallucination or generic.

On the other hand, if the response was deemed faithful, workers are asked to score three qualities: 
\textbf{Cooperativeness} means that the response is coherent with the previous turn and does not try to mislead the interlocutor or act unhelpfully. %
\textbf{Engagingness} involves engaging the interlocutor by prompting further replies and moving the conversation forward.\footnote{A low score in cooperativeness is correlated with a low score in engagingness but the opposite is not necessarily true. %
} 
\textbf{Abstractiveness} measures the ability to reuse information from the source knowledge in a novel way. 
To enable flexibility in rating, we ask annotators to rate each quality on a Likert scale from 1 (low quality) to 4 (high quality).

\paragraph{Results}
We evaluate responses generated by \textsc{T5} as it is the best performing model in terms of automated metrics (Table~\ref{tab:metrics_eval}). 
We provide human annotators with $200$ responses, where each is scored by 3 humans raters.  Results are depicted in Table \ref{tab:human_eval}. We measure the agreement for each of the 7 qualities separately using Krippendorff’s
$\alpha$ and find that the agreement (0.92, 0.91, 0.88,
0.90, 0.89, 0.75, 0.85 respectively) is reliably high.

Contrasting models trained on {\noisydata} and \newdata, we find that \newdata \ reduces hallucination by a large margin ($32.6\%$) while increasing interpretability. Also, we observe that training models on \newdata \ enhances the cooperativeness, engagingness, and abstractiveness of responses, as they tend to prompt further conversations, acknowledge previous utterances, and abstract information from the source knowledge. We see that \textsc{CTRL} benefits faithfulness but at the expense of cooperativeness and abstractiveness of the responses. The best performing model corresponds to \textsc{T5-InfoNCE}, which achieves the highest faithfulness percentage ($77.4\%$) and the highest dialogue quality scores.

\paragraph{Evaluation of unanswerable questions}
To evaluate the ability of models trained on \newdata \ to handle unanswerable questions, we analyze the responses for 200 unanswerable questions sampled from test data.
Each response is manually evaluated by 3 annotators whether the answer is appropriate.
Inter-annotator agreement based on Krippendorff’s alpha is 0.9 which is substantially high. Results indicate that \textsc{T5-InfoNCE} trained on \newdata \ substantially outperform \textsc{T5-LossTruncation} trained on \noisydata \ in answering properly unanswerable questions ($83.2\%$ vs.\ $33.3\%$).

\begin{table*}[t]
\centering
\scriptsize
\begin{tabular}{ c  l | l | c  c  c  c ||  c l l l l }
 \toprule
  
 \multirow{2}{*}{\scriptsize \textbf{Models}}
 & \multirow{2}{*}{\scriptsize \textbf{Trained on}}
   & \multirow{2}{*}{\scriptsize \textbf{Tested on}}
 &\multirow{2}{*}{\scriptsize{\textbf{Critic $\downarrow$}}} & \multicolumn{2}{c}{\scriptsize \textbf{Q$^2$} $\uparrow$}
 &{\scriptsize \textbf{F1} $\uparrow$} &
  \multirow{2}{*}{\scriptsize \textbf{Hallucination}} & 
 \multicolumn{3}{c}{\scriptsize \textbf{Faithfulness}}
 \\
  &&   & &{{\scriptsize \textbf{F1}}} &  {{\scriptsize \textbf{NLI}}} & 
  {{ $(u, \mathcal{K})$}}  &  & {{\scriptsize \textbf{Coop.}}}  &  {{\scriptsize \textbf{Abst.}}} &  {{\scriptsize \textbf{Enga.}}}\\
  \midrule
   {{\textsc{T5}}} & TopicalChat & TopicalChat & {95.0}  & {46.2} & {53.2}& {\phantom 06.6}& {\phantom 071.4\%$^{*}$} & {\textbf{3.53}} & {2.01$^{*}$} & {\bf2.56}   \\ 
     & \cellcolor{Gray}{\newdata} & \cellcolor{Gray}{TopicalChat} & \cellcolor{Gray}{\bf 59.3}  &  \cellcolor{Gray}{\bf 57.3} &  \cellcolor{Gray}{\bf 67.1}  &  \cellcolor{Gray}{\textbf{12.5}} & \cellcolor{Gray}{\bf 41.0\%} & \cellcolor{Gray}{3.07$^{*}$} & \cellcolor{Gray}{\bf 3.44} & \cellcolor{Gray}{2.20$^{*}$}  
   
  \\ 
   
  \hline

{{\textsc{T5}}} & \multirow{1}{*}{CMU-DoG} & {CMU-DoG} &  95.5  & 39.5  & 49.2 & \phantom 01.9 & {\phantom 068.4\%$^{*}$} & \bf {3.43} & {2.51$^{*}$} & {1.57$^{*}$}   \\

&  \cellcolor{Gray}{\newdata} & \cellcolor{Gray}{CMU-DoG} & \cellcolor{Gray}{\bf 21.8}&  \cellcolor{Gray}{\bf 50.5}& \cellcolor{Gray}{\textbf{57.3}}  &  \cellcolor{Gray}{\textbf{17.1}} & \cellcolor{Gray}{\bf 48.4\%} & \cellcolor{Gray}{3.29$^{*}$} & \cellcolor{Gray}{\bf 3.23} & \cellcolor{Gray}{\bf2.14} \\

\hline
{{\textsc{T5}}} & \multirow{1}{*}{\textsc{WoW}} & {\textsc{WoW}} &  57.9  & 69.4 &  72.1 &  \bf 59.6   & {48.0\%} & {2.96$^{*}$} & {1.90$^{*}$} & {1.39$^{*}$}\\

&  \cellcolor{Gray}{\newdata} & \cellcolor{Gray}{\textsc{WoW}} & \cellcolor{Gray}{\bf \phantom 07.7}&  \cellcolor{Gray}{\bf 72.9 }& \cellcolor{Gray}{\textbf{79.7}}  &  \cellcolor{Gray}{57.4}& \cellcolor{Gray}{\bf 24.2\%} & \cellcolor{Gray}{\bf 3.54} & \cellcolor{Gray}{\bf 2.67} & \cellcolor{Gray}{\bf 2.78} \\

  \bottomrule
\end{tabular}
\caption{\small Transfer results of faithful response generation from \newdata to other dialogue datasets. The most right block corresponds to human evaluation. $^{*}$ indicates that the results are statistically significant (p-value < 0.05).
}
\label{tab:transfer_learning}
\vspace{-3mm}

\end{table*}
 \subsubsection{Transfer from \newdata to other datasets}
To further examine the usefulness of \newdata \ in out-of-domain setting, we test the performance of \textsc{T5-}\newdata on TopicalChat \cite{Gopalakrishnan2019} and CMU-DoG \cite{zhou-etal-2018-dataset}, and WoW \cite{dinan2018wizard}. Contrary to \noisydata{}, speakers in CMU-DoG and TopicalChat can also take symmetric roles (i.e., both act as the wizard). Knowledge is provided from Wikipedia movie articles in CMU-DoG and from diverse sources---such as Wikipedia, Reddit and news articles---in TopicalChat.
Models are evaluated in a zero-shot setting as the corresponding training sets are not part of \newdata.
Results are depicted in Table~\ref{tab:transfer_learning}. 
Since these testing benchmarks are fraught with hallucinations (see \Cref{tab:benchamrk_hall}), we do not compare the quality of the response $u$ with respect to the gold response $g$. We report both automatic metrics and human evaluation. We follow the same human evaluation setting as before and ask 3 workers to annotate 200 responses from each model (Krippendorff’s
$\alpha$ is 0.82, 0.79, 0.85 on TopicalChat, CMU-DoG and \noisydata respectively). 
In this regard, the models trained on \newdata are far more faithful than the models trained on in-domain data despite the distribution shift. 
For example, \textsc{T5-FaithDial}  tested on TopicalChat test data decreases hallucination by $35.7$ points on Critic, by $13.9$ points on Q$^2$-NLI and by $30.4$ points on human scores. 
Similar trends can be observed for \textsc{TopicalChat} and \noisydata (except for F1 on WoW, yet human evaluation shows humans prefer \newdata models by a large margin of $23.8$). Generated responses can be found in Table~\ref{tab:benchmark:wow_begin_dist}.
Regarding other dialogue aspects, \textsc{T5-FaithDial} models tested on TopicalChat and CMU-DoG enjoy a larger degree of abstractiveness than in-domain models but have lower scores of cooperativeness and engagingness. However, all of these aspects are enhanced when tested in-domain on WoW.

\section{Related Work}

\paragraph{Hallucination in Natural Language Generation.}
Hallucination in knowledge-grounded neural language generation has recently received increasing attention from the NLP community \cite{ji2022survey}. Tasks include data-to-text generation \cite{wiseman-etal-2017-challenges, parikh-etal-2020-totto},  machine translation  \cite{raunak-etal-2021-curious,wang-sennrich-2020-exposure}, summarization \cite{durmus-etal-2020-feqa,kang-hashimoto-2020-improved}, generative question answering \cite{li2021addressing} and dialogue generation \cite{dziri-etal-2021-neural, dziri2021evaluating, rashkin-etal-2021-increasing}.

These works focus on either devising automatic metrics to identify when hallucination occurs
\citep{wiseman-etal-2017-challenges}
or finding possible causes for this degenerate behaviour, including out-of-domain generalization and noisy training data points \citep{kang-hashimoto-2020-improved,raunak-etal-2021-curious} and exposure bias caused by MLE training \citep{wang-sennrich-2020-exposure}.

\paragraph{Hallucination in Dialogue Systems.}
Hallucinations in knowledge-grounded neural dialogue generation is an emergent research problem \cite{roller-etal-2021-recipes,mielke2020linguistic,shuster-etal-2021-retrieval-augmentation,dziri-etal-2021-neural, rashkin-etal-2021-increasing}.
Existing work aims predominantly to address hallucinations via engineering loss functions or enforcing consistency constraints, for instance by conditioning generation on control tokens
\citep{rashkin-etal-2021-increasing}, by learning a token-level hallucination critic to flag problematic entities and replace them \citep{dziri-etal-2021-neural}, or by augmenting the dialogue system with a module retrieving relevant knowledge \citep{shuster-etal-2021-retrieval-augmentation}.

Although promising, these approaches are prone to replicate---or even amplify---the noise found in training data.
\citet{dziri-etal-2022-origin} demonstrated that
more than $60\%$ of three popular dialogue benchmarks are rife with hallucination, which is picked up even by models designed to increase faithfulness.
To the best of our knowledge, \newdata{} is the first dataset for information-seeking dialogue that provides highly faithful curated data. %

\paragraph{Hallucination Evaluation.}
Recently introduced benchmarks can serve as testbeds for knowledge grounding in dialogue systems, such as BEGIN  \citep{dziri2021evaluating}, DialFact \cite{gupta2021dialfact}, Conv-FEVER \cite{santhanam2021rome} and Attributable to Identified Sources (AIS) framework \cite{rashkin2021measuring}.
Meanwhile, a recent study has reopened the question of the most reliable metric for automatic evaluation of hallucination-free models, with the Q$^2$ metric \cite{honovich-etal-2021-q2} showing performance comparable to human annotation. In this work, we further contribute to this problem by proposing a critic model---trained on our collected {\critic} data---that achieves high performance on 
the  \BEGIN{} benchmark.

\section{Conclusions}

We release \newdata{}, a new benchmark for faithful information-seeking dialogue, where a domain-expert bot answers queries based on gold-standard knowledge in a conversational manner.
Examples are created by manually editing hallucinated and uncooperative responses in Wizard of Wikipedia (\noisydata{}), which constitute 79.1\% of the original dataset. Leveraging the resulting high-quality data, we train both a hallucination critic, which discriminates whether utterances are faithful to the knowledge and achieves a new state of the art on \BEGIN{}, and several dialogue generation models. In particular, we propose strategies to take advantage of both noisy and cleaned data, such as intermediate fine-tuning on \noisydata{} and an auxiliary contrastive objective. With both automated metrics and human evaluation, we verify that models trained on \newdata{} drastically enhance faithfulness and abstractiveness, both in-domain and during zero-shot transfer to other datasets, such as TopicalChat and CMU-DoG.

 \section*{Acknowledgements}
We are grateful to the anonymous reviewers for
helpful comments. We would like to thank MTurk workers for contributing to the creation of \newdata and for giving feedback on various pilot rounds. 
SR acknowledges the support of the IBM-Mila
grant, the NSERC Discovery grant and the Facebook CIFAR AI chair program. OZ acknowledges
the Alberta Machine Intelligence Institute Fellow
Program and the Canadian Institute for Advanced
Research AI Chair Program.

\clearpage
\bibliography{anthology, custom}
\appendix 
\section{AMT Instructions}
\label{huma_ann_amt}
Here, we detail the instructions given to workers in the annotation task. We follow instructions from \cite{dziri-etal-2022-origin}  in determining \BEGIN and \vrm categories. And, according to the identified categories, we ask workers to perform a particular edit. Below are the questions we ask in every HIT:
\begin{enumerate}
    \item Does the \expert's response contain other information that is NOT supported by $\mathcal{K}$? (e.g., facts, opinions, feelings) (Yes/No)
     \begin{enumerate}
         \item If the response is hallucinated, what is the type of the unsupported information? 
         (options: expressing a personal experience, expressing an opinion, expressing feelings, expressing unsupported facts, giving advice, acknowledging information from the \textsc{Seeker})
         \item  If the response is hallucinated, was the unsupported information triggered by a question/opinion from the \textsc{Seeker}? (Yes/No)
         \item Besides unsupported information, does the \expert's response contain thoughts/opinions/feelings/facts that are supported by $\mathcal{K}$? (Yes/No)
         \item Modify the \expert's sentence such that the response:
          \begin{enumerate}
        \item uses only the facts from $\mathcal{K}$ to make the response informative.
\item is not a copy paste of $\mathcal{K}$ but a paraphrase of it.
\item is relevant to the previous utterance and cooperative with the \textsc{Seeker}.
            \end{enumerate}
            
             \item If the response is not hallucinated, does the \expert's response express personal thoughts/opinions/feelings that are supported by $\mathcal{K}$? (Yes/No)
        \item If the response is not hallucinated, does the \expert's response contain factual/objective information that is supported by $\mathcal{K}$? (Yes/No)
 \end{enumerate}
   
        \item If the answer is ``No" to (e) and (f), the response is flagged as generic. We ask the annotators to modify the \expert's sentence such that the response is supported by $\mathcal{K}$.
        \item If the response is  faithful, workers are asked the following question: Is the \expert's response cooperative with the \textsc{Seeker}'s response? i.e. the \expert does not ignore answering a question, or does not act in any unhelpful way. 
        \begin{enumerate}
            \item If yes, no modification is required for the \expert's response.
            \item If no, modify the bot sentence such that:
            \begin{enumerate}
                \item The response is relevant to the previous utterance and cooperative with the \textsc{Seeker}.
                \item The response is not a copy paste of $\mathcal{K}$ but a paraphrase of it.
            \end{enumerate}
        \end{enumerate}
    
\end{enumerate}

\section{Pay Structure}
\label{appendix:pay}
 We pay crowdworkers a base pay of $\$1.7$/HIT (USD). To retain excellent workers for all rounds, we give a bonus of $\$35-\$40$ per 100 HITs that are submitted successfully. The average amount of time spent per HIT is 6 min, i.e., in one hour, workers are able to complete $10$ HITS. This is equivalent to $\$17-\$18$ per hour.

\section{Abstractiveness strategies}
\label{app:abstraction}
We annotate manually 150 responses to explore the techniques used by the \expert \ to derive and represent information from the knowledge source $\mathcal{K}$. Table~\ref{tab:abstraction}  shows the different abstractiveness types with their frequencies:
\paragraph{Inference} corresponds to information which can be derived from the evidence with an intermediate step in reasoning; in other words, it involves inferring obvious but implicit information from $\mathcal{K}$, from the Apprentice utterance, or from commonsense knowledge. It encompasses \textit{implicatures} (e.g. replace ``She finished some of her work" with ``She did not finish all of her work"), \textit{presuppositions} (e.g. replace ``She stopped smoking" with ``She used to smoke"), and \textit{deductions} (e.g. replace ``She drove her car to work every day for 3 years" with ``She can drive"). Also, it includes \textit{commonsense knowledge} (e.g. replace ``Elvis, the artist, \dots" with ``Elvis, a person, \dots").

\paragraph{Rewording} involves the replacement of words/phrases in $\mathcal{K}$ with similar wording. One instance of Rewording is \textit{synonymization}, where words/phrases are replaced with their synonyms (e.g. replace ``can lead to" with ``can result in"). Also, it is sometimes possible to preserve truth while replacing words/phrases denoting subset members with their supersets, as in \textit{generalization} (e.g. replace ``Some dogs" with ``Some animals"), or superset members with their subsets, as in \textit{specification} (e.g. replace ``all animals" with ``all dogs"). Lastly, \textit{pronominalization} replaces pronouns with noun phrases, or vice versa (e.g. , replace ``Andy visited Mary" with “Andy visited her”).

\paragraph{Restructuring} corresponds to restructuring the syntactic formulations (\textit{syntax}) of $\mathcal{K}$ in a meaning-preserving manner. It can be done through passivization (e.g. replace ``Andy visited Mary" with ``Mary was visited by Andy"). Another type of Restructuring is \textit{reordering}, the rearranging of list elements. \textit{Ellipsis} refers to the ellipsis of sentences or the expanding of ellipted sentences (e.g.  replace ``I have not heard of Elvis" with ``I have not"). \textit{Questioning} refers to the restructuring of declarative statements into questions.

\paragraph{Abridging} refers to the removal of  modifiers and/or optional complements while preserving the entailment relationship between $\mathcal{K}$ and the response. This includes removing  \textit{adjectives}, \textit{adverbs}, and \textit{independent clauses} (e.g. replace ``I’m taking the red bus early today, in 10 minutes" with ``I'm taking the bus today").

\paragraph{Bridging} involves adding words/phrases to connect or introduce parts of the utterance (e.g.  ``So…", ``In other words, \dots", ``In addition, \dots", etc).

\begin{table*}[t]
\centering
\small
\resizebox{\textwidth}{!}{
\begin{tabular}{ L{5cm} | L{6cm} | L{5cm} | c}
 \toprule
{\textbf{Abstractiveness Type}}
 &{ \textbf{Knowledge}}
   & {\textbf{Response}}
 &{\%}  \\
  \midrule
  \vspace*{-1.5\baselineskip}
  \begin{itemize}[leftmargin=*]
    \item[] Abridging
    \item[] Rewording 
        \vspace*{-.5\baselineskip}
        \begin{itemize}[leftmargin=*]
            \item[] 
            \textbf{Pronominalization}, \underline{Synonymization}
        \end{itemize}
    \item[] Restructuring
        \vspace*{-.5\baselineskip}
        \begin{itemize}[leftmargin=*]
            \item[] \footnotesize{\textcolor{blue}{Questioning}, \textcolor{violet}{Reordering}}
        \end{itemize}
  \end{itemize}\vspace{-\baselineskip}\mbox{} &
   The name `Track and Field' \underline{is derived from} \textbf{the sport's} typical venue:  \textcolor{violet}{a stadium with an oval running track \underline{enclosing} a grass field where the throwing and jumping events take place}.
    &
   \textcolor{blue}{Did you know that} the name Track and Field \underline{comes from} \textbf{its} typical venue?  \textcolor{violet}{That's a stadium with a grass field \underline{inside} of it and a running track}. & 25.99
  \\
  
 \hline
 \vspace*{-1.5\baselineskip}
    \begin{itemize}[leftmargin=*]
        \item[] Abridging
        \item[] Restructuring
        \vspace*{-.5\baselineskip}
        \begin{itemize}[leftmargin=*]
            \item[] \footnotesize{Syntax}
        \end{itemize}
        \item[] Inferring
            \vspace*{-.5\baselineskip}
            \begin{itemize}[leftmargin=*]
                \item[] \footnotesize{\textcolor{red}{Commonsense}}, \footnotesize{\textcolor{teal}{Implicature}}
            \end{itemize}
    \end{itemize}\vspace{-\baselineskip}\mbox{} &
  \textcolor{red}{The earliest precursor to the modern rap} is the \textcolor{red}{West African griot tradition}, in which ``oral historians'', or ``praise-singers'', would disseminate oral traditions and genealogies, or use their formidable rhetorical techniques for gossip or to ``praise or critique individuals.''
    &
\textcolor{teal}{I don't have any specific recommendations for rap artists}. \textcolor{red}{I do have some history.} Rap's precursor is called \textcolor{red}{griot, from West Africa}. & 12.6
  \\
  \hline
    \vspace*{-1.5\baselineskip}
    \begin{itemize}[leftmargin=*]
        \item[] Abridging
        \item[] Restructuring
        \vspace*{-.5\baselineskip}
        \begin{itemize}[leftmargin=*]
            \item[] \footnotesize{\textcolor{olive}{Syntax}}
        \end{itemize}
    \end{itemize}\vspace{-\baselineskip}\mbox{} &
 Among other licensing requirements, \textcolor{olive}{different countries require} pharmacists to hold either a Bachelor of Pharmacy, Master of Pharmacy, or Doctor of Pharmacy degree.
    &
I can also tell you that degrees like Bachelor of Pharmacy and Doctor or Master of Pharmacy can be \textcolor{olive}{required by different countries}. & 7.33
  \\

 \hline
 \vspace*{-1.5\baselineskip}
 \begin{itemize}[leftmargin=*]
    \item[] Abridging
    \item[] Rewording
    \vspace*{-.5\baselineskip}
    \begin{itemize}[leftmargin=*]
        \item[]  \textbf{Pronominalization}, \underline{Synonymization}
    \end{itemize}
    \item[] Inferring
    \vspace*{-.5\baselineskip}
    \begin{itemize}[leftmargin=*]
        \item[]\textcolor{red}{Commonsense}, \footnotesize{\textcolor{teal}{Implicature}}
    \end{itemize}
 \end{itemize}\vspace{-\baselineskip}\mbox{} &
  Use by a \underline{wider audience} only \textcolor{teal}{came in 1995 when} \textcolor{red}{restrictions} on the use of the \textbf{Internet} to carry commercial traffic were lifted.
    &
 \underline{More people} \textcolor{teal}{started using} \textbf{it} \textcolor{teal}{after} \textcolor{red}{some restrictions} on internet use were lifted in 1995.& 8.66
  \\



 \hline
 \vspace*{-1.5\baselineskip}
    \begin{itemize}[leftmargin=*]
        \item[] Inferring
            \vspace*{-.5\baselineskip}
            \begin{itemize}[leftmargin=*]
                \item[] \footnotesize{\textcolor{brown}{Deduction}}
            \end{itemize}
    \end{itemize}\vspace{-\baselineskip}\mbox{} &
 Homebrewing is the brewing of beer \textcolor{brown}{on a small scale for personal, non-commercial purposes}.
    &
Interesting that you've done homebrewing before. So you just brew \textcolor{brown}{enough for yourself}? & 4.6
  \\
  \bottomrule
\end{tabular}}
\caption{\small Possible abstractiveness strategies of {\newdata} from manual analysis on 200 responses. 
}

\label{tab:abstraction}
\vspace{-5mm}
\end{table*}

\section{Implementation Details}
\label{sec:implement}
\paragraph{Critic} We implement all our critics using the Huggingface Transformers library \cite{wolf-etal-2020-transformers}. We train all models for $10$ epochs, using  a batch size of $32$ and the Adam optimizer \cite{kingma2015adam} with a learning rate of $1 \times 10^{-5}$. We warm up the learning rate for $6\%$ of the training steps followed by a linear decay. 

\paragraph{Generation Models} We implement all the baselines using the Huggingface Transformers library \cite{wolf-etal-2020-transformers} and the Pytorch-lightning library\footnote{\url{https://github.com/PyTorchLightning/pytorch-lightning}}. We train our models for 10 epochs on a batch size of $32$ via accumulating gradients for 4 steps, and use Adam with a learning rate of $6.25 \times 10^{-5}$ that warms up for 4\% of training steps, followed by a linear decay. The models are evaluated twice per epoch on the validation set, and the best performing model is saved for testing. We early stop the training with a patience of 5. The maximum dialogue history length is set to $3$ utterances.
For DoHA \cite{prabhumoye-etal-2021-focused}, we follow the same hyperparameters, used in the paper. More specifically, DoHA is trained for 25 epochs using an Adam optimizer with a learning rate of $2 \times 10^{-5}$, a warm-up ratio of 0.1, and accumulating gradients for 8 steps. For CTRL, the code is not publicly available. We were able to reproduce the results ourselves by following training implementations in the paper and exchanging discussions with the authors.
Training for all models is done on an Nvidia V100 GPU 32GB and for inference, we use nucleus sampling with p=0.6. 





\begin{table*}[ht]
\footnotesize
    \centering
    \begin{tabular}{ L{2cm}  l L{11cm}}
    \toprule
    \textbf{Data}  & \textbf{Dialogues} \\
    \midrule
    {\bf TopicalChat}   &
         {\bf Knowledge} & Football is a family of team sports that involve, to varying degrees, kicking a ball to score a goal.   Unqualified, the word football is understood to refer to whichever form of football is the most popular in the regional context in which the word appears. Sports commonly called football in certain places include association football ( known as soccer in some countries); gridiron football (specifically American football or Canadian football); Australian rules football; rugby football (either rugby league or rugby union); and Gaelic football. These different variations of football are known as football codes. \textbf{The University of Iowa's locker room for visiting football teams is completely painted pink}. In the 1960's top bowlers made twice as much as top football stars. \textcolor{teal}{The highest score ever in a football game occurred in 1916} when Georgia Tech defeated Cumberl and 222-0 ESPN won an emmy for the creation of the superimposed yellow line representing the first down line for American football games. Former Partiots RB BenJarvus Green - Ellis has never fumbled the football in his NFL career. \\
        &
       {\bf History}&Did you know that the University of Iowa 's locker room is painted pink? I wonder why? \\
        &
        \textbf{T5-\newdata} &  \textcolor{teal}{I don't know honestly. A room is painted pink? that's interesting. Speaking of football, did you know that the highest score in football appeared in 1916?} (\underline{Faithful})\\
        &
          \textbf{T5-TopicalChat} &  \textcolor{red}{That is pretty funny, I guess the visiting team gets a lot of attention from them, I think they are scared of the pink locker room.} (\underline{Hallucination})
        \\ \midrule
        \textbf{CMU-DoG} & 
        Knowledge: & Cast: Lindsay Lohan as Cady Heron, Rachel McAdams as Regina George, Lacey Chabert as Gretchen Wieners, Amanda Seyfried as Karen Smith, Lizzy Caplan as Janis Ian. critical response: Boasts a one -two- three punch in star Lindsay Lohan, screenwriter Tina Fey and director Mark Waters, and, indeed, it delivers a knockout The screenplay was highly praised by critics with Peter Travers of Rolling Stone calling it 'comic gold'. In November 2012, Rotten Tomatoes named it in its 'Top 50 Greatest Teen Comedies'. Director: Mark Waters.   \textcolor{teal}{Genre: Comedy}, teeny introduction: \textcolor{teal}{Mean Girls is a 2004 American teen comedy film directed by Mark Waters and written by Tina Fey}. \textcolor{teal}{The film is partially based on Rosalind Wiseman's 2002 non-fiction self-help book Queen Bees and Wannabes,} which describes female high school social cliques and the damaging effects they can have on girls. MovieName: Mean Girls rating: Rotten Tomatoes: 84\% and average: 6.9/10. Metacritic Score: 66/100. CinemaScore :  \textcolor{teal}{A year: 2004}. \\
        & History: & I think Rachel McAdams had an even 
  better role as Regina George however ! \\
        & \textbf{T5-\newdata}: & \textcolor{teal}{That's great. Did you know it was written by Tina Fey? That's being said, the movie is based on a 2002 book called Queen Bees and Wannabes about social cliques in high school? It's also a comedy film. It's a 2004 film, directed by Mark Waters.} (\underline{Faithful})\\ 
        & \textbf{T5-CMU-DoG}: & \textcolor{red}{I think it was really good. I bet it will win the Oscars.} (\underline{Halluciantion}) \\

        \bottomrule
    \end{tabular}
    \caption{Examples from \textsc{T5-\newdata} tested on out-of-domain data: TopicalChat and CMU-DoG.}
    \label{tab:benchmark:wow_begin_dist}
\end{table*}

\bibliographystyle{acl_natbib}

\end{document}